\pdfoutput=1

\documentclass[11pt]{article}

\usepackage{EMNLP2022}

\usepackage{times}
\usepackage{latexsym}

\usepackage[T1]{fontenc}

\usepackage[utf8]{inputenc}

\usepackage{microtype}

\usepackage{inconsolata}

\title{Unifying Discrete Reasoning over Table and Text \\ with Logical Programming}
\title{{Unified Discrete Reasoning over Table and Text as Program Generation}}
\title{UniRPG: Unified Discrete Reasoning over Table and Text \\as Program Generation}

\author{Yongwei Zhou\textsuperscript{1}, Junwei Bao\textsuperscript{2}\thanks{~~Work was done during the first author’s internship at JD AI mentored by Junwei Bao: baojunwei001@gmail.com.}, Chaoqun Duan\textsuperscript{2}, Youzheng Wu\textsuperscript{2}, \\ \bf Xiaodong He\textsuperscript{2}, Tiejun Zhao\textsuperscript{1}\\
    \textsuperscript{1}Harbin Institute of Technology \;\;  \textsuperscript{2}JD AI Research \\
    ywzhou@hit-mtlab.net \;\;  baojunwei001@gmail.com  \;\; tjzhao@hit.edu.cn
    }

\usepackage{bibentry}

\usepackage[utf8]{inputenc}
\usepackage{subfigure}
\usepackage{booktabs}
\usepackage{amsmath}
\usepackage{amsfonts}
\usepackage{multirow}
\usepackage{color}
\usepackage{enumitem}
\usepackage{microtype}
\usepackage{tabularx}

\usepackage{chngpage}
\usepackage{array}
\usepackage{microtype}
\usepackage{xcolor}
\usepackage{fontenc}
\usepackage{arydshln}
\usepackage{array}
\usepackage{booktabs}

\usepackage{newfloat}
\usepackage{listings}
\usepackage{makecell}
\usepackage{aliascnt}

\usepackage{algorithm}
\usepackage{algorithmic}
\usepackage{hyperref}
\usepackage{graphicx}
\usepackage{amssymb}

\newcommand{\tabincell}[2]{\begin{tabular}{@{}#1@{}}#2\end{tabular}}
\usepackage{ulem}

\usepackage[utf8]{inputenc}
\usepackage{cleveref}
\crefname{section}{§}{§§}
\Crefname{section}{§}{§§}

\begin{document}
\maketitle

\begin{abstract} 
Question answering requiring discrete reasoning, e.g., arithmetic computing, comparison, and counting, over knowledge is a challenging task.
In this paper, we propose \textbf{UniRPG}, a semantic-parsing-based approach advanced in interpretability and scalability, to perform \textbf{Un}ified d\textbf{i}screte \textbf{R}easoning over heterogeneous knowledge resources, i.e., table and text, as \textbf{P}rogram \textbf{G}eneration. 
Concretely, UniRPG consists of a neural programmer and a symbolic program executor,
where a program is the composition of a set of pre-defined general atomic and higher-order operations and arguments extracted from table and text.
First, the programmer parses a question into a program by generating operations and copying arguments, and then the executor derives answers from table and text based on the program.
To alleviate the costly program annotation issue, we design a distant supervision approach for programmer learning, where pseudo programs are automatically constructed without annotated derivations.
Extensive experiments on the TAT-QA dataset show that UniRPG achieves tremendous improvements and enhances interpretability and scalability compared with state-of-the-art methods, even without derivation annotation.
Moreover, it achieves promising performance on the textual dataset DROP without derivations.
\footnote{The code is released at \url{https://github.com/JD-AI-Research-NLP/UniRPG}.}

\end{abstract}
\section{Introduction}

\begin{figure}[t]
    \centering
    \includegraphics[width=3in]{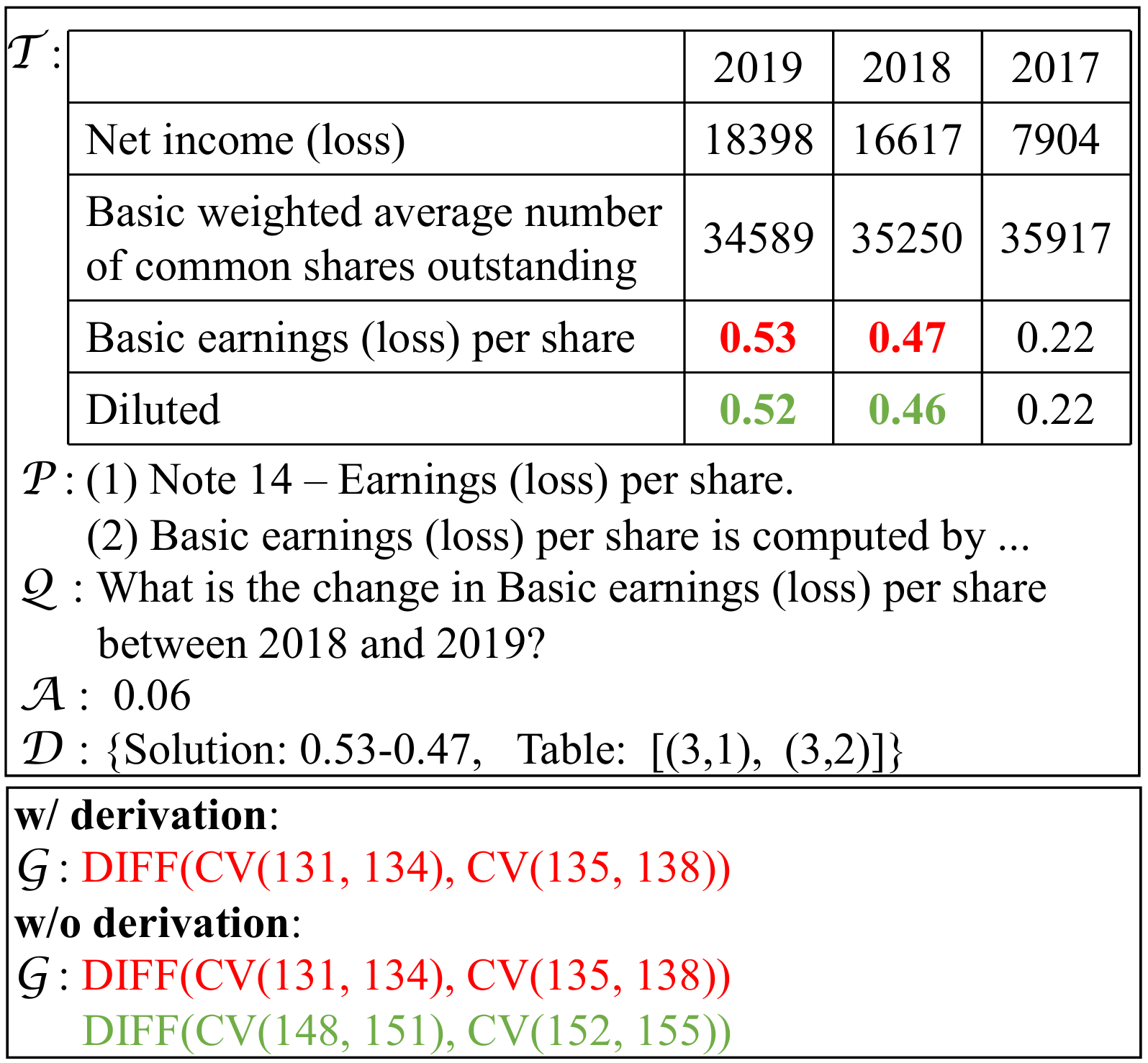}
    \caption{The upper block shows a training instance from the TAT-QA dataset. It consists of a table $\mathcal{T}$, question $\mathcal{Q}$, paragraphs $\mathcal{P}$, answer $\mathcal{A}$ and derivation $\mathcal{D}$.
    The bottom block displays the constructed programs $\mathcal{G}$ under w/ and w/o derivation settings.
    The gold program is in red, and the others are in green.
    }
    \label{fig:program_example}
    \vspace{-0.3cm}
\end{figure}

Question answering requiring discrete reasoning, e.g., arithmetic computing, sorting, comparison, and counting, over knowledge is a challenging task and has attracted wide attention.
Advanced approaches to this task are roughly divided into mixture-of-experts (MoE)-based and semantic parsing (SP)-based methods.
The former divides questions into several limited categories according to their required reasoning type and designs multiple specialized predictors to deal with each category~\cite{dua2019drop,ran2019numnet,chen2020question,zhu2021tat,Zhou2022OPERAOD}.
The latter translates questions into programs such as SPARQL queries~\cite{bao2014knowledge,bao-etal-2016-constraint,chen2019neural,gan2021towards,cao2021lgesql,hui-etal-2022-s2sql}, to derive the answers. 
Compared with MoE-based methods, SP-based methods have advantages in interpretability and scalability, and we mainly explore discrete reasoning along this technical route in this paper. 

As to works based on SP-based methods, they always research over homogeneous {knowledge resource}, {i.e., either} unstructured text~\cite{dua2019drop,ran2019numnet,chen2020question,Zhou2022OPERAOD} or {structured tables and knowledge graphs}~\cite{pasupat2015compositional,herzig2020tapas,liu2021tapex}.  
{Respectively, they design a specialized semantic parser for each knowledge category according to the corresponding data characteristic.}
However, {in practical scenarios} (e.g., financial reports and sports news analysis),
it usually requires a model to comprehensively understand heterogeneous {knowledge resources} (e.g., tables and text) and perform reasoning over them.
{Therefore, to extend SP-based methods to more general and realistic scenarios, it needs a unified framework for discrete reasoning over heterogeneous knowledge resources.}
Meanwhile, to mitigate the {program} annotation cost, it also needs an automatic approach to produce pseudo programs.

To address the above challenges, we propose {\textbf{UniRPG}, a SP-based approach}, to perform \textbf{Un}ified d\textbf{i}screte \textbf{R}easoning over table and text as \textbf{P}rogram \textbf{G}eneration.
Specifically, UniRPG consists of a neural programmer and a deterministic symbolic program executor, {where a program is the composition of a set of pre-defined general atomic and higher-order operations and arguments extracted from table and text}.
{The programmer leverages a structure-aware knowledge reader to encode heterogeneous knowledge, and a program generator to decode programs through generating operations and copying arguments from table and text.
During inference, decoding constraints are leveraged to ensure program legality.}
{Then, the executor derives answers from table and text based on the program.
To alleviate the costly program annotation issue, we design a distant supervision approach for programmer learning, where pseudo programs are automatically constructed without annotated derivations.
Furthermore, a re-weight strategy is introduced to measure the importance of pseudo programs for noise reduction in training.
}

To verify the effectiveness and generality of UniRPG, we conduct comprehensive experiments on the TAT-QA~\cite{zhu2021tat} and DROP~\cite{dua2019drop} datasets under w/ and w/o derivation settings.
Experiment results show that UniRPG achieves tremendous improvements and enhances interpretability and scalability compared with previous state-of-the-art (SOTA) methods on the TAT-QA dataset.
Specifically, it outperforms the previous SOTA model by 17.0 EM and 18.0 F1 on the test set under full supervision.
Besides, our weakly supervised model based on automatically constructed pseudo-programs only lags behind fully supervised model 1.0 EM and 1.4 F1 points.
Moreover, it also achieves promising performance on the textual dataset DROP in weak supervision.
In conclusion, the contributions of this work are summarized as follows:
(1) We propose an effective program-generation-based framework for discrete reasoning over table and text;
(2) To alleviate costly program annotation, we design a distant supervision approach to construct pseudo programs automatically;
(3) We conduct comprehensive experiments to verify the effectiveness and scalability of UniRPG on the TAT-QA and DROP datasets.

\begin{table*}[t]
\resizebox{1\textwidth}{!}{
\begin{tabular}{ l c c}
\hline
   \bf {Operations} & \bf{Arguments} & \bf{Descriptions} \\ \hline
\tabincell{l}{\textbf{Atomic Operations}\\ $\mathtt{SPAN}$ \\ $\mathtt{CELL}$ \\ $\mathtt{VALUE}$ \\$\mathtt{CELL\_VALUE}$ ($\mathtt{CV}$) } & 
\tabincell{c}{\\ $(s, e)$ \\ $(s, e)$ \\ $(s, e)$ \\ $(s, e)$} & 
\tabincell{c}{\\ Extract a piece of text or a number from table and \\ paragraphs based on the index of start(s) and end(e) tokens. } \\ \hline

\tabincell{l}{\textbf{Higher-order Operations}} & & \\

\tabincell{l}{$\mathtt{KV}$} & \tabincell{c}{$\langle \mathtt{CELL}, \mathtt{CV} \rangle$, $\langle \mathtt{SPAN}, \mathtt{VALUE} \rangle$}& \tabincell{c}{Build a key-value pair with a text piece and a number.} \\

\tabincell{l}{$\mathtt{COUNT}$} & \tabincell{c}{Atomic Operations}& \tabincell{c}{Return the number of extracted text pieces/numbers. } \\

\tabincell{l}{$\mathtt{MULTI\_SPANS}$} & \tabincell{c}{Atomic Operations}& \tabincell{c}{Return all the extracted text pieces/numbers.} \\

\tabincell{l}{$\mathtt{ARGMAX/ARGMIN}$} & \tabincell{c}{$\mathtt{KV}$} & 
\tabincell{c}{Return the key with maximum/minimum value \\ from all the key-value pair candidates.} \\

\hdashline 

\tabincell{l}{$\mathtt{SUM}$/$\mathtt{DIFF}$/$\mathtt{TIMES}$/$\mathtt{DIV}$ } & 
\tabincell{c}{$\mathtt{VALUE}$, $\mathtt{CV}$, Constants} &
\tabincell{c}{Compute the sum/difference/product/quotient \\ between two numbers.} \\

\tabincell{l}{$\mathtt{AVG}$ \\ $\mathtt{CHANGE\_R}$} & 
\tabincell{c}{$\mathtt{VALUE}$, $\mathtt{CV}$, Constants \\ $\mathtt{VALUE}$, $\mathtt{CV}$, Constants} &
\tabincell{c}{ Compute the average value of argument numbers. \\ Compute the rate of change between two numbers.} \\ \hline

\tabincell{l}{\textbf{Constants}\\ $\mathtt{0}$ / $\mathtt{1}$ / $\mathtt{100}$} & \tabincell{c}{\\ -} & \tabincell{c}{ \\ Several constants are commonly involved in math problems.} \\ \hline

\end{tabular} 
}
\caption{\label{table:statics} \label{table:Operations} All the defined operations, their arguments and descriptions.}
\vspace{-0.3 cm}
\end{table*}

\section{Methodology\label{approach_section}}
\subsection{Task Definition}
The task of discrete reasoning over hybrid data aims to predict the answer from the given input (e.g., tables and text). 
Formally, given a question $\mathcal{Q}$, a table $\mathcal{T}$, and a set of paragraphs $\mathcal{P}$, the target is to derive the answer $\mathcal{A}^{\ast}$ from $\mathcal{T}$ and $\mathcal{P}$.
\begin{align}
    \mathcal{A}^{\ast} = \arg \max P(\mathcal{A}|\mathcal{Q},\mathcal{T},  \mathcal{P};\theta).
    \label{Eq:task}
\end{align}
Specifically, the table $\mathcal{T}$ consists of $m \times n$ cells $\{c_{ij}\}$, where $m$ and $n$ are the number of rows and columns.
The paragraph set $\mathcal{P} = \{p_1, p_2, ... , p_k\}$ contains $k$ paragraphs which are relevant to the table.
Notably, 
the answer $\mathcal{A}^{\ast}$ in this task is either a span extracted from the table and the paragraphs or a number computed based on them.
Therefore, it requires the model to perform discrete reasoning such as arithmetic operations, counting, comparison, sorting, and their compositions.

\subsection{Framework Overview}
As mentioned above, the method of this task needs to be capable of performing discrete reasoning over heterogeneous data.
To this end, we first define a set of operations as reasoning units, which consists of atomic and high-order operations.
Based on these operations, we propose to perform unified discrete reasoning over table and text as program generation (UniRPG).
Concretely, UniRPG consists of a neural programmer and a symbolic program executor, and it can be formulated as:
\begin{align}
    \begin{split}
        &P(\mathcal{A}|\mathcal{Q},\mathcal{T}, \mathcal{P};\theta) \\
        = &\sum_{\mathcal{G}}{\mathbb{I}(f(\mathcal{G})=\mathcal{A})P(\mathcal{G}|\mathcal{Q},\mathcal{T},\mathcal{P};\theta)},
    \end{split}
    \label{Eq:framework}
\end{align}
where $P(\mathcal{G}|\mathcal{Q},\mathcal{T},  \mathcal{P};\theta)$ denotes the \textit{neural programmer} and $f(\mathcal{\cdot}) $ is the symbolic \textit{program executor}.
$\mathbb{I}(f(\mathcal{G})=\mathcal{A})$ is an indicator function with value 1, if the answer $\mathcal{A}$ can be derived based on the program $\mathcal{G}$, and 0 otherwise.

\subsection{Operations}
As shown in Table~\ref{table:Operations}, we define 15 operations, which consist of 4 atomic operations and 11 high-order operations.
The atomic operations indicate extracting a piece of text or a number from the table or paragraphs.
Specifically, $\mathtt{SPAN}$ and $\mathtt{VALUE}$ are used to extract content from paragraphs while $\mathtt{CELL}$ and $\mathtt{CELL\_VALUE (CV)}$ are used to extract content from the table.
The atomic operations are the foundation for unified discrete reasoning over table and text.
As to high-order operations, they indicate how to process the arguments extracted based on atomic operations.
For example, $\mathtt{KV}$, $\mathtt{ARGMAX}$, and $\mathtt{ARGMIN}$ are tailored for
questions requiring comparison or sorting.
$\mathtt{COUNT}$ and $\mathtt{MULTI\_SPANS}$ are used to solve counting and multi-spans extraction problems, respectively.
We also introduce some arithmetic operations to deal with math questions, including $\mathtt{SUM}$, $\mathtt{DIFF}$, $\mathtt{TIMES}$, $\mathtt{DIV}$, $\mathtt{AVG}$ and $\mathtt{CHANGE\_R}$.
Therefore, we formulate the priority of atomic operations higher than high-order operations.
In addition, we also propose some constants that are commonly used in math problems.

\subsection{Neural Programmer}
In this work, we adopt the BART~\cite{lewis2020bart} to implement the Neural Programmer.
To adapt it to this task, we have improved its encoder and decoder respectively.
\begin{figure*}[th]
    \centering
    \includegraphics[width=6.3in]{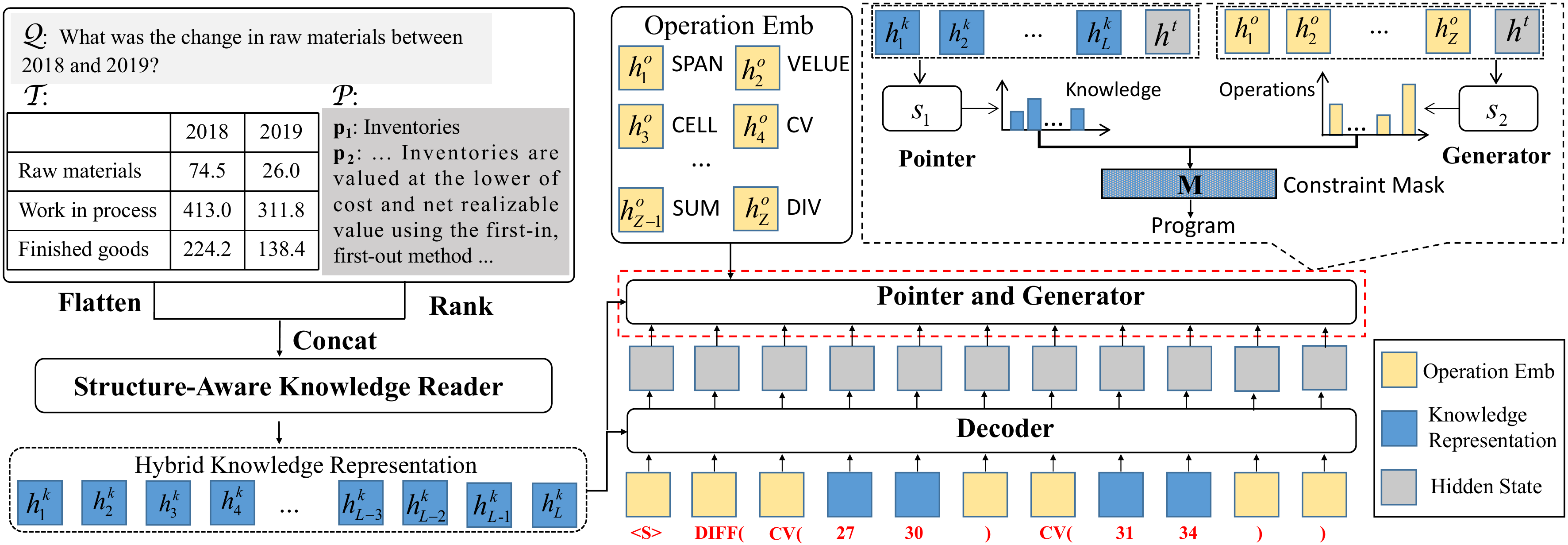}
    \caption{The architecture of neural programmer. It consists of a structure-aware knowledge reader and a program generator. 
    $\bf{M}$ is a decoding constraint mask to ensure program legality.}
    \label{fig:model}
\end{figure*}

\subsubsection{Structure-Aware Knowledge Reader}
The original encoder of BART only takes textual data as input.
However, in this task, besides paragraphs, the input contains tables as well.
Following previous works~\cite{liu2021tapex,zhou2022table,zhang2020table}, we first flatten the table into a sequence and concatenate it with the paragraphs together, and then feed the processed paragraphs and table into the encoder.
Formally, we rank paragraphs based on the similarity between each one with the question.
As to the table, we flatten it by rows and 
then construct the input as follows:
\begin{align}
    \mathcal{S} = [\mathtt{\langle s \rangle}; \mathcal{Q} ;\mathtt{\langle /s \rangle}; \mathcal{T}; \mathtt{\langle /s \rangle}; \mathcal{P}; \mathtt{\langle /s \rangle}].
\end{align}
Then, we feed it into the encoder to learn the hybrid knowledge representation $\mathbf{H}^{k} = \{\mathbf{h}_i^{k}\}_{i=1}^{L}$,
where $L$ is the length of the input sequence.

Unlike the text that only contains the linear structure, tables have a more complex 2D structure.
Therefore, simple concatenation can not completely capture the structural information in the table.
Inspired by~\citet{zhang2020table}, we inject table structure information into self-attention layers with a structure-aware attention mask.
Specifically, in the lower layers, each cell only takes information from other cells in the same row, question, and paragraphs into account.
In the upper layers, the information of cells in the same column is also integrated into the representation of the target cell.
This way, the encoder can learn the lexical features in the lower layers and capture complex cross-row reasoning in the upper layers.
Finally, we denote this encoder as a structure-aware knowledge reader.

\subsubsection{Programmer Generator}
In this work, the program generator is responsible for generating programs that are used to perform discrete reasoning over table and text.
Specifically, an executable program contains several operations.
For the atomic operations, they need to take the start and end indexes of the content in the text and the table as input.
For the high-order operations, they take the output of the atomic operations as input.
Therefore, the generator should decode the program based on operations, text, and table.
However, the original decoder of the BART~\cite{lewis2020bart} is used to generate sentences.
To bridge the gap, we substitute the prediction vocabulary with the operation set and integrate the copy mechanism~\cite{gu-etal-2016-incorporating,see2017get} into the BART to retrieve reasoning-related numbers or text pieces from table and text as operation arguments when decoding programs.

Formally, we extra define a trainable embedding matrix $\{\mathbf{h}^{o}_j\}_{j=1}^{Z}$ to indicate the operations and the $j$-th row corresponds to the $j$-th operation $o_j \in \mathcal{O}$.
Based on the decoder of the BART, we can compute two sets of scores at each step as follows:
\begin{align}
    \label{eq:similarity}
    s^{k, t}_i &= s_{1}(\mathbf{h}^t, \mathbf{h}_i^{k}),  i \in [1, 2, 3, ..., L], \\
    s^{o, t}_j &= s_{2}(\mathbf{h}^t, \mathbf{h}^{o}_j),   j\in [1, 2, ...,  Z],
\end{align}
where $s^{k, t}_i$ and $s^{o, t}_j$ denote the scores of the $i$-th token of the input and $j$-th operation.
$\mathbf{h}^t$ is the hidden state of the decoder at the $t$-th step.
$\mathbf{h}_i^{k}$ is the contextual representation of the $i$-th token in the input.
$s_1(\cdot)$ and $s_2(\cdot)$ are both cosine similarity functions.
$s_1(\cdot)$ is a pointer to locate the copied numbers and text pieces from the input sequence $\mathcal{S}$.
$s_2(\cdot)$ acts as a generator to select an operation from the operation set $\mathcal{O}$. 
After obtaining these two scores, we normalize them with a softmax function.

\subsection{Program Executor}
In this work, the program is executable, and we can directly apply it to perform discrete reasoning over table and text.
Therefore, we implement a symbolic program executor rather than a specific neural module. 
Given a program, we can execute it based on the meaning of each operation in it.
For example, given a program $\mathtt{SUM(CV(s_1, e_1), CV(s_2, e_2))}$, since the priority of atomic operations is higher than high-order operations, $\mathtt{CV(s_1, e_1)}$ and $\mathtt{CV(s_2, e_2)}$ are first executed to extract value from the table.
And then, $\mathtt{SUM(\cdot)}$ is executed to return the summation of extracted value from the table as the output.

\section{Training and Inference}
\label{sec:training and inference}

\subsection{Training}
\label{subsec:training_instance_construction}

\paragraph{Training Instance Construction: {w/ Derivation}}
Each training instance $\mathcal{I} = \{\mathcal{Q}, \mathcal{T}, \mathcal{P}, \mathcal{A}, \mathcal{D}\}$ consists of a question $\mathcal{Q}$, a table $\mathcal{T}$, associated paragraphs $\mathcal{P}$, an answer $\mathcal{A}$ and a derivation $\mathcal{D}$.
The derivation $\mathcal{D}$ indicates the answer {position in table and text} and the {reasoning} solution to derive the answer.
We can deterministically construct a program based on {such a training instance $\mathcal{I}$ that contains annotated derivation $\mathcal{D}$}.
As the example shown in Figure \ref{fig:program_example}, we determine the unique program $\mathcal{G} = \mathtt{DIFF(CV(131,134), CV(135,138))}$
based on the {reasoning} solution $0.06=0.53-0.47$ and the {positions} of these two numbers.

\paragraph{Training Instance Construction: {w/o Derivation}}
In addition, to alleviate the costly derivation annotation, we design a distant supervision method to construct pseudo programs automatically without derivation.
We design a template library $\mathcal{B}$ shown in Table~\ref{table:templates} in the Appendix~\ref{app:templates}, which consists of five categories, namely text/number extraction, multi-spans extraction, counting, comparison/sorting, and arithmetic computing.
We search all the possible programs for each training instance based on $\{\mathcal{Q}, \mathcal{T}, \mathcal{P}, \mathcal{A}, \mathcal{B}\}$.
For text/number extraction, we find all the occurrences of answers from table and text.
Concerning multi-spans extraction, we find all the occurrences of each span and then combine them.
For counting questions, building a fine-grained counting program fails due to a lack of the {positions} of answer clues.
To address the issue, we convert a question with a multi-spans answer into a counting question by replacing its interrogatives (e.g. ``What'', ``Which'', ``Who'') with ``How many'', and leverage its answer items to build the counting programs.
For questions requiring comparison/sorting, we first find the answer position in the table and construct key-value pairs with cells in the same row/column and cells in other rows/columns that contain values.
If performing the $\mathtt{ARGMAX/ARGMIN}$ operation on the list of key-value pairs yields the answer, we then append it to the program candidates.
Additionally, we define several math templates that conduct specific operations on numbers in table and text for math problems.
Afterward, for each training instance in TAT-QA, {more programs (eight on average) that can derive correct answers are found without using annotated derivation $\mathcal{D}$}.
{For instance, another program $\mathtt{DIFF}(\mathtt{CV}(131,134),\mathtt{CV}(135,138))$ is found under w/o derivation setting shown in Figure \ref{fig:program_example}.}
Moreover, based on the templates, we find at least one pseudo program for 89\% training instances in the TAT-QA dataset~\cite{zhu2021tat}.

\paragraph{Training Objective} 
We consider all the programs as supervision constructed under the w/ derivation setting.
{For w/o derivation setting, we introduce} a re-weight strategy to measure the importance of pseudo programs for noise reduction in training. 
Formally, the program generation {loss $\mathcal{L}_{g}$ is defined} as follows:
\begin{equation}
\label{eq:loss}
\mathcal{L}_{g} = - \sum_{\mathcal{G} \in \Omega} \alpha_{\mathcal{G}}\log p(\mathcal{G}|\mathcal{Q},\mathcal{T},\mathcal{P};\theta).
\end{equation}
{Precisely, we calculate a weight $\alpha_{\mathcal{G}}$ for each pseudo program $\mathcal{G} \in \Omega$ as the reciprocal of the number of pseudo programs with the same operations.
Based on the weight distribution, losses of pseudo programs are weighted summed to treat each type of program fairly.}
In addition, TAT-QA has another task to predict the scale of answer and it may be
\textit{None}, \textit{Thousand}, \textit{Million}, \textit{Billion} and \textit{Percent}.
A correct final answer requires that both the predicted answer text and scale are correct.
We propose a 5-categories classifier for scale prediction and joint train it with the neural programmer.
Denote the cross-entropy loss for scale prediction as $\mathcal{L}_{s}$ and then the total loss can be calculated with a weight $\lambda$ as $\mathcal{L} = \mathcal{L}_{g} + \lambda\mathcal{L}_{s}$.

\subsection{Inference with Constraint Decoding} 
To ensure the legality of the program,
we propose the following three categories of decoding constraints and utilize a decoding mask to filter the illegal program candidates at each step. 
(1) \textit{Index constraints}. For example, with $(s,e)$ as arguments to an atomic operation, the end index $e$ must be equal to or larger than the start index $s$. 
(2) \textit{Type constraints}. 
For example, the operation $\mathtt{CV}$ should return a number from the table rather than paragraphs.  
(3) \textit{Composition constraints}. 
Each operation must take the correct type of operations as arguments. 
More decoding constraint details are shown in Appendix \ref{sec:decoding_constraints}.

\section{Experiment}
\subsection{Dataset and Evaluation Metrics}

The experiments are conducted on two QA datasets, including TAT-QA~\cite{zhu2021tat} and DROP~\cite{dua2019drop}.
TAT-QA is a large-scale QA dataset in finance created to facilitate discrete reasoning over table and text.
DROP is a large-scale QA dataset requiring discrete reasoning over text. 
It is crowd-sourced based on passages from Wikipedia.
The TAT-QA dataset especially provides the derivation for each training instance, but the DROP dataset does not.
The statics information of the two datasets is summarized in Table~\ref{table:data_statics}. 
We employ Exact Match (EM) and F1 as the evaluation metrics.

\begin{table}[t]
\centering
\resizebox{0.46\textwidth}{!}{
\begin{tabular}{|l|c|l|l|l|l|}
\hline
    \bf{Dataset} & \bf{Derivation} & \bf{Train} & \bf{Dev} & \bf{Test} & \bf{Total} \\ \hline
\tabincell{l}{\textbf{TAT-QA} \\ \#question \\ \#hybrid contexts } & \tabincell{c}{Yes} & \tabincell{c}{ \\ 13215 \\ 2201 } & \tabincell{c}{ \\ 1668 \\ 278 } & \tabincell{c}{ \\ 1669 \\ 278 } &  \tabincell{c}{ \\ 16552 \\ 2757 }\\ \hline
 \tabincell{l}{\textbf{DROP} \\ \#question \\ \#passage} & {No} & \tabincell{c}{ \\ 77409 \\ 5565 } &  \tabincell{c}{ \\ 9536 \\ 582 } &   \tabincell{c}{ \\ 9622 \\ 588 } & \tabincell{c}{\\96567 \\ 6735}\\ \hline
\end{tabular} }
\caption{\label{table:data_statics} The statics of the two datasets. \# means the number of each item.}
\vspace{-0.6cm}
\end{table}

\subsection{Implementation Details}

Note that the implementation details are described in the Appendix~\ref{app:implement details}. 

\subsection{Baselines}
\paragraph{Baselines for TAT-QA}
(1) \textit{Textual QA Models:} BERT-RC~\cite{devlin2019bert} is a simple extractive reading comprehension (RC) model that predicts a text span as the answer.
NumNet+v2 ~\cite{ran2019numnet} leverages a graph neural network to perform reasoning over numbers in tables and paragraphs.
When applying RC models to TAT-QA, the table is flattened into a sequence by row and concatenated with questions and paragraphs.
(2) \textit{Tabular QA Models:} The baseline TaPas for WTQ employs TaPas~\cite{herzig2020tapas} as the knowledge encoder.
TaPas contains prior knowledge by pre-training over large-scale tabular data and injects table structure information in the embedding layer.
(3) \textit{Hybrid QA Models:} HyBrider ~\cite{chen2020hybridqa} first links the question with a table cell and feeds paragraphs and the selected cell into the RC module to infer the answer.
TAGOP~\cite{zhu2021tat} is an MoE-based model that defines multiple operations as predictors to support discrete reasoning.
It first detects the evidence from tables and paragraphs and selects an answer predictor to infer the answer based on the evidence.

\paragraph{Baselines for DROP}
NAQANet~\cite{dua2019drop} designs three specific modules to handle different types of questions, including span extraction, counting, and arithmetic calculation.
MTMSN~\cite{hu2019multi} introduces an extra answer predictor to address negation questions and a re-ranking mechanism to rank arithmetic expression candidates considering their context information.
NumNet~\cite{ran2019numnet} builds a bi-directional fully-connected graph to model the magnitude of numbers and employs a graph neural network to reason over it.
GenBERT~\cite{geva2020injecting} injects numerical reasoning capability into BERT by pre-training with large-scale synthetic numerical data.

\begin{table}[t]
  \small
  \centering
  \resizebox{0.49 \textwidth}{!}{
  \setlength{\tabcolsep}{1mm}{
	\begin{tabular}{l  cccc}
		\toprule
		\multirow{2}*{\bf Method}     &\multicolumn{2}{c}{\bf Dev } &   \multicolumn{2}{c}{\bf Test}\\ 
		 \cmidrule(lr){2-3} \cmidrule(lr){4-5}
		&\bf EM &\bf F1  &  \bf EM & \bf F1 \\ 
		\midrule
        \textbf{Baselines (w/ derivation)} \\ 
		BERT-RC~\cite{devlin2019bert} & 9.5  & 17.9 & 9.1 & 18.7 \\ 
		NumNet+v2~\cite{ran2019numnet} & 38.1 & 48.3 & 37.0 & 46.9 \\ 
        TaPas for WTQ~\cite{herzig2020tapas} & 18.9 & 26.5 & 16.6 & 22.8 \\ 
        HyBrider~\cite{chen2020hybridqa} & 6.6 & 8.3 & 6.3 & 7.5 \\
        TAGOP~\cite{zhu2021tat}  & 55.2 & 62.7 & 50.1 & 58.0 \\
        \midrule
        \textbf{w/o derivation} \\
         UniRPG(BART-base)  & 66.1 & 73.4 & 61.5 & 70.8 \\
        UniRPG(BART-large) & \bf 69.1 & \bf 76.4 & \bf 66.1 & \bf 74.6 \\
        \textbf{w/ derivation} \\ 
        UniRPG(BART-base)  & 67.5  & 74.8 & 64.4  & 73.6 \\
        UniRPG(BART-large) & \bf 70.2 & \bf 77.9 & \bf 67.1 & \bf 76.0 \\
		\midrule
		Human  & & & 84.1 & 90.8\\
		\bottomrule
	\end{tabular}
	} 
	}
	\caption{\label{table:results} Results of baselines and our models on the TAT-QA dataset. 
	}
\end{table}

\begin{table}[t]
  \small
  \centering
  \resizebox{0.49 \textwidth}{!}{
  \setlength{\tabcolsep}{1mm}{
	\begin{tabular}{l  cccc}
		\toprule
		\multirow{2}*{\bf Method}     &\multicolumn{2}{c}{\bf Dev } &   \multicolumn{2}{c}{\bf Test}\\ 
		 \cmidrule(lr){2-3} \cmidrule(lr){4-5}
		&\bf EM &\bf F1  &  \bf EM & \bf F1 \\ 
		\midrule
		NAQANet~\cite{dua2019drop}  & 46.20 &49.24 &44.07 &47.01\\
		NumNet~\cite{ran2019numnet}  & 64.92 &68.31 &64.56 &67.97\\
	    GenBERT~\cite{geva2020injecting}  & 68.80 & 72.30&68.60 &72.35 \\
	    MTMSN~\cite{hu2019multi}  &76.68 & 80.54&  75.88& 79.99\\
    	UniRPG & \bf{78.02} & \bf{80.93} & \bf 77.08 & \bf 80.42  \\
		\bottomrule
	\end{tabular}
	} }
	\caption{\label{table:results_drop} Results on the DROP dataset.}
	\vspace{-0.5cm}
\end{table}

\subsection{Main Results}
\label{sec:main_results}

\begin{figure}[t]
  \centering
	\includegraphics[width=3.0in]{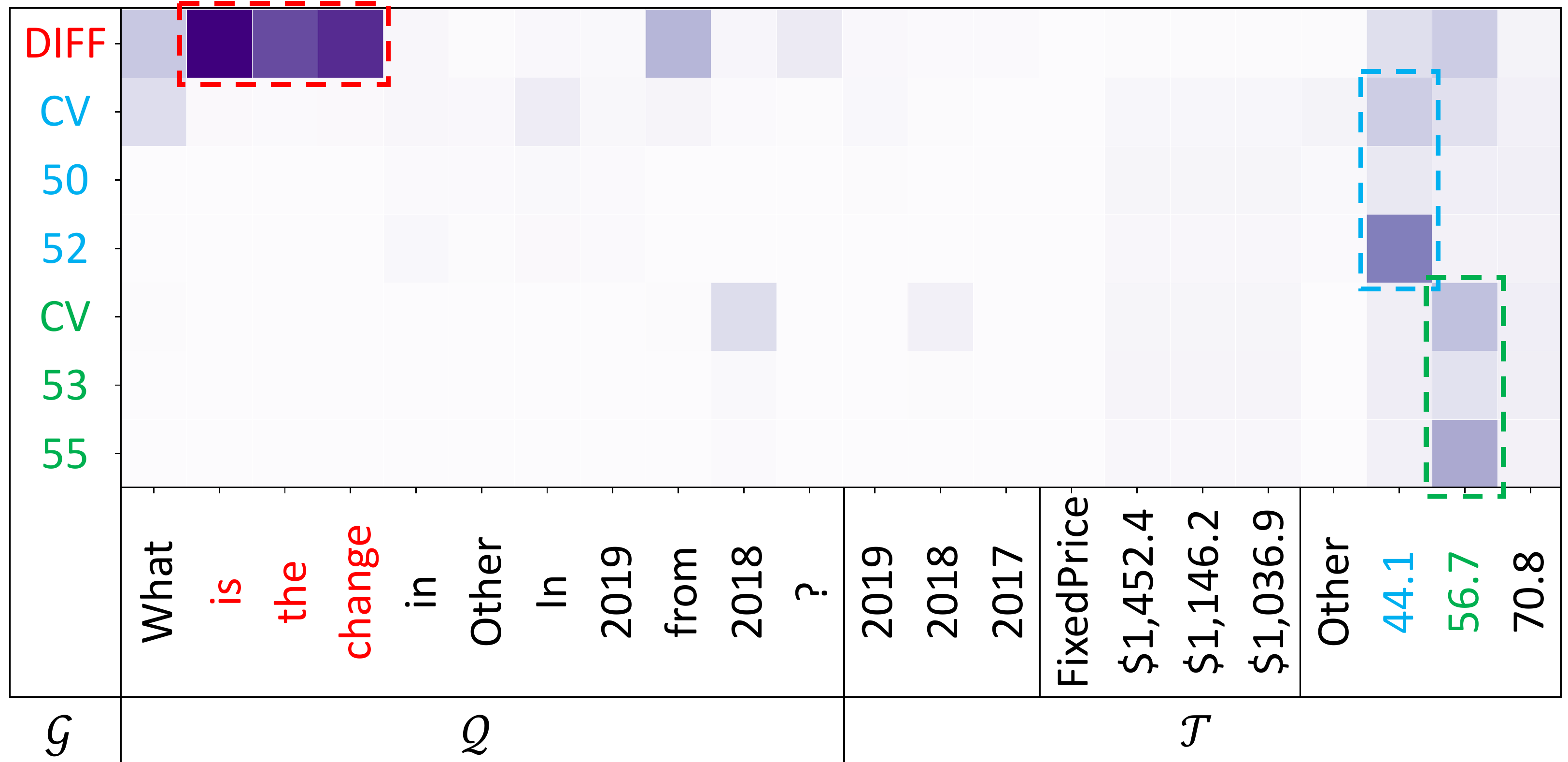}
	\caption{The visualization of the attention score in the last encode layer.
	The horizontal axis represents the token sequence of the question and table.
    The vertical axis represents decoded program units at different time steps.
    Darker color means a higher attention score.}
	\label{Fig:visualization}
\end{figure}

\paragraph{Result on TAT-QA } Table \ref{table:results} displays the experiment results of baselines and UniRPG on the TAT-QA dataset.
All the baseline results come from previous work~\cite{zhu2021tat}.
We implement UniRPG based on BART~\cite{lewis2020bart} and train it under the w/ and w/o derivation settings, respectively.
As shown in Table~\ref{table:results},
UniRPG(BART-base) and UniRPG(BART-large) respectively exceed the previous SOTA method TAGOP by 14.3 EM/15.6 F1 and 17.0 EM/18.0 F1 on the test set under full supervision (w/ derivation).
In addition, our framework also works well under weak supervision (w/o derivation), exceeding all the baselines by a remarkable margin.
Moreover,  our weakly-supervised model based on automatically constructed pseudo
programs only lags behind fully-supervised models 1.0 EM and 1.4 F1 on the test set.
We think the promising improvement comes from the following aspects:
(1) UniRPG can solve more complex and more types of questions, such as compositional reasoning and abstract answers (answers do not appear in context).
(2) As visualized in Figure~\ref{Fig:visualization}, UniRPG captures the relationship between programs and hybrid knowledge.  

Moreover, Table~\ref{table:detailed_results} displays the detailed experiment results of baseline (TAGOP) and our UniRPG(BART-large) under w/ derivation setting on the test set w.r.t answer types and sources.
Our method significantly outperforms TAGOP, whether the answers come from table, text, and collaboration of table and text.
Furthermore, UniRPG has advantages in the following three types of answers, including Span, Spans, and Arithmetic answers,  yet disadvantages in Counting answers. 
We speculate the phenomenon stems from the lack of training instances with Counting answers, which only account for 2\%.

\paragraph{Results on DROP}
We verify the effectiveness of UniRPG over text based on the DROP dataset. 
In this setting, unlike reasoning over table and text, UniRPG does not need to consider some operations tailored for table and some non-involved reasoning types, such as $\mathtt{CELL}$, $\mathtt{CELL\_VALUE}$ and $\mathtt{TIMES}$/$\mathtt{DIV}$. 
As shown in Table~\ref{table:results_drop}, UniRPG also achieves promising performance on the textual dataset DROP with 77.08 EM and 80.42 F1 scores.
Although the performance of UniRPG does not exceed previous SOTA models tailored for DROP, it provides a unified framework for reasoning over single data and hybrid data types.

\begin{table}[t]
  \small
  \centering
  \resizebox{0.49 \textwidth}{!}{
  \setlength{\tabcolsep}{1mm}{
	\begin{tabular}{l cccc}
		\toprule
		 &  \bf{Table} &  \textbf{Text} & \textbf{Table-Text} & \textbf{Total}\\ 
		\midrule
		Span &   57.8 | 67.6    &   70.6 | 79.5     &  71.7 | 85.4  &   67.9 | \bf 78.0 \\
        Spans &  77.0 | 84.1     &  59.1 | 60.2    &  76.9 | 78.9     &  75.1 | \bf 78.5\\
        Counting &   63.6 | 18.1  &  - | -         & 62.1 | 31.0       & \textbf{62.5} | 27.0 \\
        Arithmetic &  41.1 | 75.8 &  27.3 | 36.3  & 46.5 | 76.7     & 42.5 | \bf 75.3\\
        Total   &  49.3 | \bf 73.8   &  68.7 | \bf 76.5       &  62.2 | \bf 77.9    &   58.0 | \bf 76.0 \\
		\bottomrule
	\end{tabular}
	} }
	\caption{\label{table:detailed_results} The detailed experiment results (F1 score) of TAGOP and our model under w/ derivation setting w.r.t. answer types and sources. The left side of the delimiter ``|'' is the results of TAGOP, and the right is ours.
	}
\end{table}

\begin{table}[t]
  \small
  \centering
  \resizebox{0.49 \textwidth}{!}{
  \setlength{\tabcolsep}{1mm}{
	\begin{tabular}{lcccccc}
		\toprule
		 \multirow{2}*{\bf Settings}     &  \multicolumn{2}{c}{\bf BART-base} &  \multicolumn{2}{c}{\bf BART-large} & \multicolumn{2}{c}{\bf Average Gain}\\ 
		 \cmidrule(lr){2-3} \cmidrule(lr){4-5} \cmidrule(lr){6-7}
		& \bf EM & \bf F1  &  \bf EM & \bf F1  &   \bf EM & \bf F1 \\ 
		  \midrule
		  \bf{w/ derivation} \\
		  UniRPG & \bf67.5 & \bf 74.8 & \bf 70.2 & \bf77.9  & \\
		  ~~~~ w/o constraints &  65.2 & 73.7 & 68.5 & 76.8 & 2.0~$\downarrow$ & 1.1~$\downarrow $\\
		  ~~~~ w/o structure & 63.1 & 70.8 & 69.6&77.4 & 2.5~$\downarrow$ & 2.2~$\downarrow$\\
		  \midrule
		 \bf{w/o derivation} \\
		  UniRPG & \bf 66.1 & \bf 73.4 & \bf 69.1 & \bf 76.4 & \\
		  ~~~~ w/o constraints & 64.9 &  72.3  & 67.8 & 75.0 & 1.2 $\downarrow$ & 1.2 $\downarrow$\\
		  ~~~~ w/o re-weight  & 63.7  & 71.1 & 67.5 & 75.1 & 2.0 $\downarrow$ & 1.8 $\downarrow$\\        
		  ~~~~ w/o structure & 62.2 & 69.1 & 64.5 & 72.3 & 4.2 $\downarrow$ & 4.2 $\downarrow$ \\
    	\bottomrule 
	\end{tabular} 
	} }
	\caption{\label{table:ablation}  Ablation results on the development set of the TAT-QA dataset.}
\end{table}

\subsection{Ablation Study}
We investigate the effect of decoding constraints, modeling of table structure, and re-weight strategy for noise reduction in training on the development set of the TAT-QA dataset.
As shown in Table~\ref{table:ablation}, the performance declines by 2.0 EM/1.1 F1 and 1.2 EM/1.2 F1 on average under w/o and w/ derivation settings when removing the decoding constraints (w/o constraints).
It demonstrates that constraint decoding can significantly reduce errors caused by illegal programs.
Moreover, if removing the modeling of table structure, the performance declines by 2.5 EM/2.2 F1 and 4.2 EM/4.2 F1 on average under w/ and w/o derivation settings, respectively.
It indicates that modeling table structure significantly contributes to the performance of our models. 
Additionally, we explore the influence of the reweighted mechanism for pseudo programs under the w/o derivation setting.
Table \ref{table:ablation} shows the re-weight strategy effectively improves the performance by 2.0 EM and 1.8 F1 on average, which treats each type of pseudo program fairly and alleviates negative impact of noise in training.

\begin{figure}[t]
  \centering
	\includegraphics[width=2.8in]{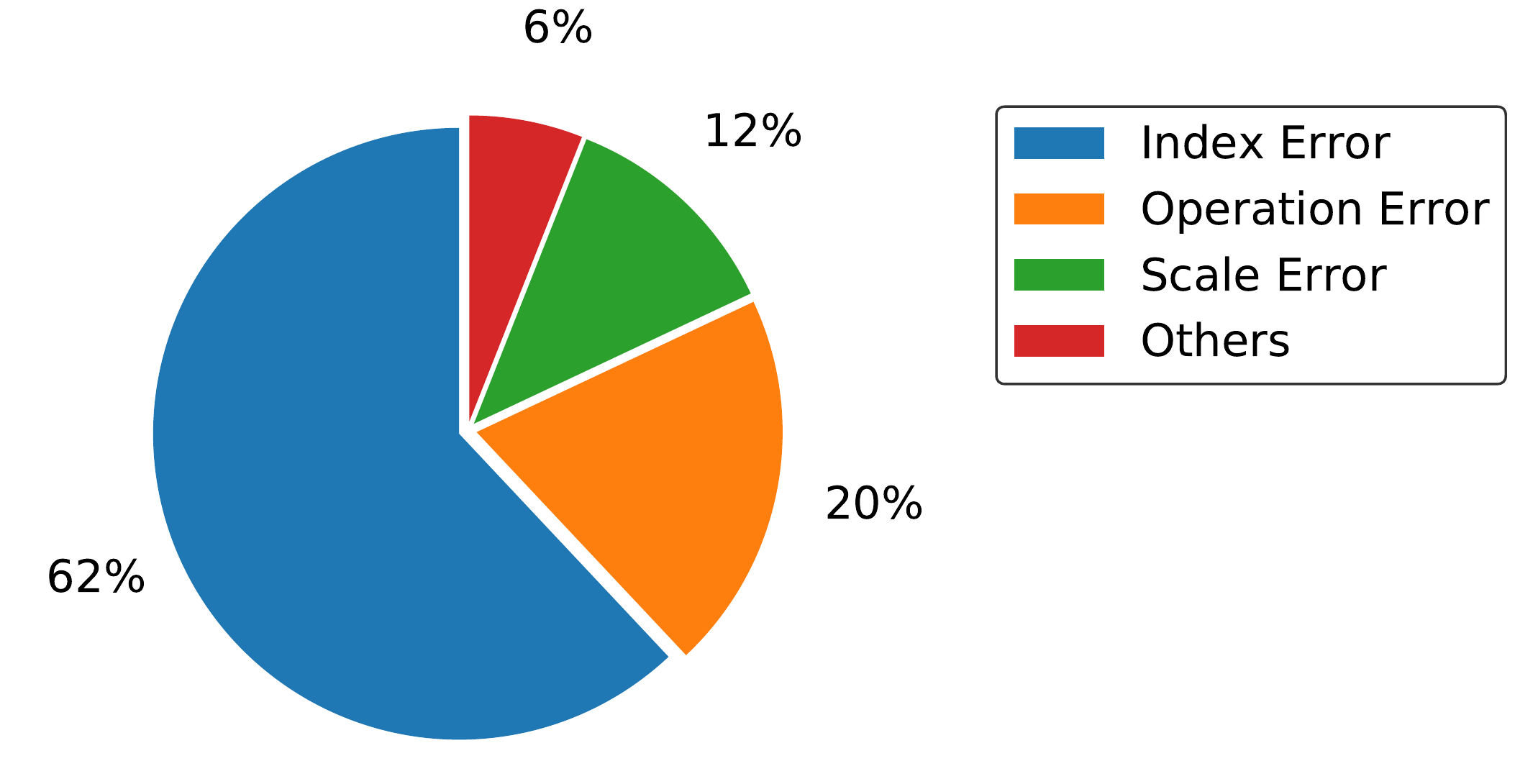}
	\caption{\label{op_statics} The error distribution of our model.}
\end{figure}

\section{Discussion}
\paragraph{Interpretability Analysis}
As shown in Figure~\ref{Fig:program_examples}-\ref{Fig:program_examples_3}, our model generates fine-grained programs that intuitively demonstrate the discrete reasoning process.
Furthermore, we visually investigate how the neural programmer decodes programs based on questions and hybrid knowledge resources.
Figure \ref{Fig:visualization} displays the cross-attention score distribution from the last layer of the decoder based on a random sample.
The horizontal axis is the input sequence $\mathcal{S}$, and the vertical axis represents the decoded program units at different time steps.
We observe that cross-attention visualization reflects the underlying relationships between the question prefix ``\textit{What is the change}'' and operation $\mathtt{DIFF}$.
Moreover, $\mathtt{CV}$ correctly extracts the two numbers 44.1 and 56.7 from the table.

\paragraph{Error Analysis}
We analyze our error predictions on the dev set of TAT-QA and observe the error instances fall into the following four categories.
The first category is that atomic operations extract wrong text pieces or numbers due to index error, accounting for 62\%.
We think it is intractable to derive the start and end index from the entire input sequence precisely, which typically has hundreds of tokens.
The second category is caused by operation errors when decoding programs and accounts for 20\%.
We speculate that this error stems from some questions being inherently confusing.
For example, given a question ``\textit{What is the average USD-EUR exchange rate in FY 2019?}'',  the programmer wrongly predicts the operation $\mathtt{AVG}$ to solve the question since the keyword \textit{average}
appears in the question. 
However, the correct annotated derivation is ``70.07/80.82'', i.e., $\mathtt{DIV}$ is the proper operation.
The remaining errors come from the scale prediction and others, which account for 
18\%.
In summary, more accurately extracting text pieces and numbers from table and text is essential to further improve our model.

\begin{figure}[t]
  \centering
	\includegraphics[width=2.6in]{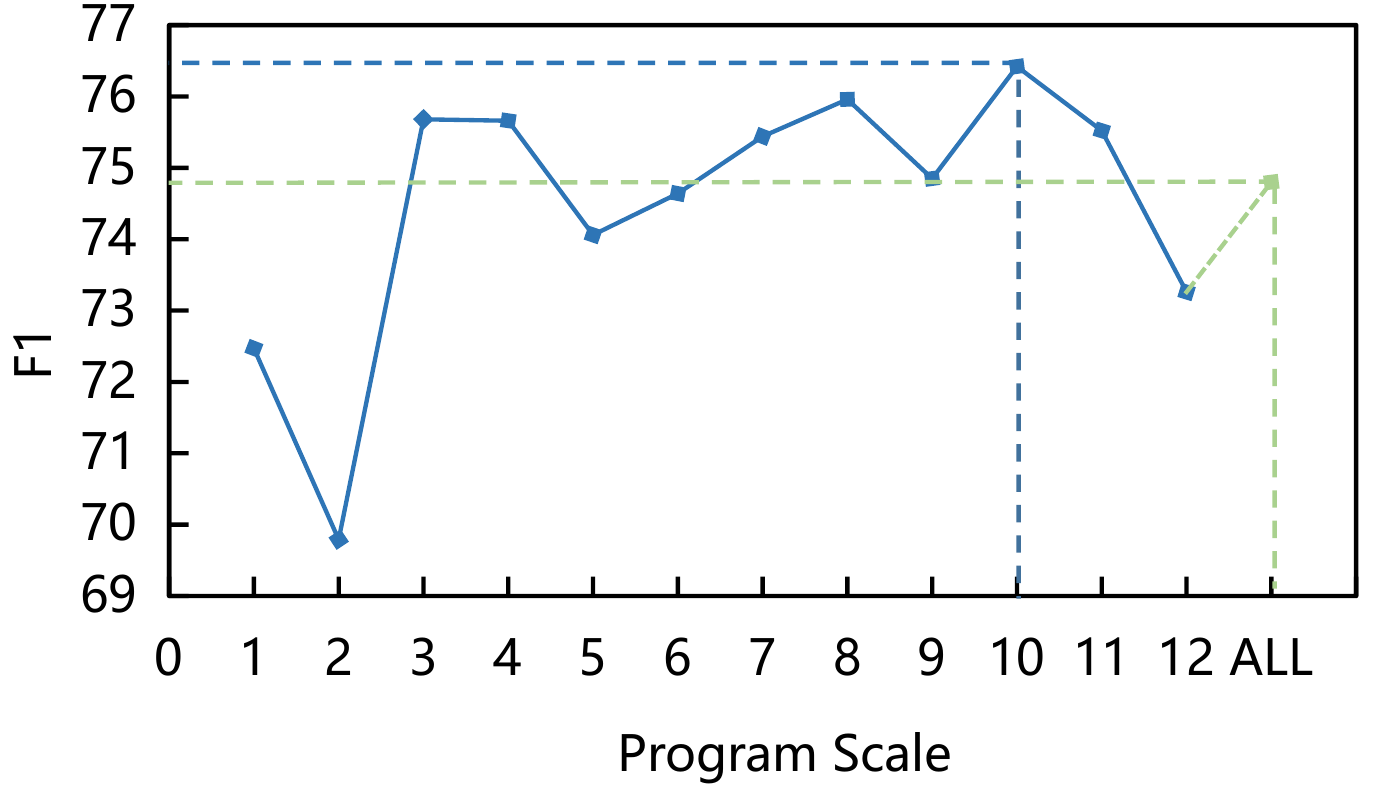}
	\caption{The influence of program scale. \texttt{ALL} means training with all the pseudo programs.}
	\label{Fig:program_scale}
	\vspace{-0.5cm}
\end{figure}

\paragraph{Influence of Program Scale}
For each training instance in TAT-QA,  there are 8 pseudo programs to derive answers correctly on average under the w/o derivation setting.
To explore the influence of the program scale, we respectively train models with Top $N$ programs of each instance and all the programs, where $N=1,2,..., 12$. 
Figure~\ref{Fig:program_scale} displays the F1 curve of our models on the development set of TAT-QA with respect to the scale of programs $N$.
The F1 score increases with the program scale, then shows a relatively stable performance, reaching a peak when $N=10$.
It indicates the initial increase of F1 is due to gradually sufficient training data, and our model shows good robustness as the noisy data increases.

\section{Case Study}
We compare our UniRPG with the baseline TAGOP on 50 instances randomly sampled from the development set of TAT-QA, where TAGOP predicts wrongly, but our model predicts correctly.
UniRPG has advantages in the following three aspects compared with TAGOP.
(1) Capture of argument order.
TAGOP introduces a 2-classifier to learn the argument order with a one-hot order label, ignoring the modeling of the interaction between two arguments. 
Our framework can naturally learn the parameter order of operations because program sequences are inherently ordered.
As shown in example (a), our methods can avoid 60\% error instances of TAGOP~\cite{zhu2021tat} due to the wrong argument order.
(2) Evidence detection.
TAGOP extracts evidence by classifying each token of the input and then predicts answers based on the evidence, which is prone to accumulating errors.
We employ a copy mechanism to extract content from the input, which only needs to predict the start and end tokens.
Example (b) shows our methods reduce 20\% error predictions caused by inaccurate evidence compared with TAGOP.
(3) Compositional reasoning.
As shown in example (c), our model has better scalability that can perform compositional with multiple operations compared to TAGOP.

\begin{figure*}[t]
    \centering
    \includegraphics[width=6.3in]{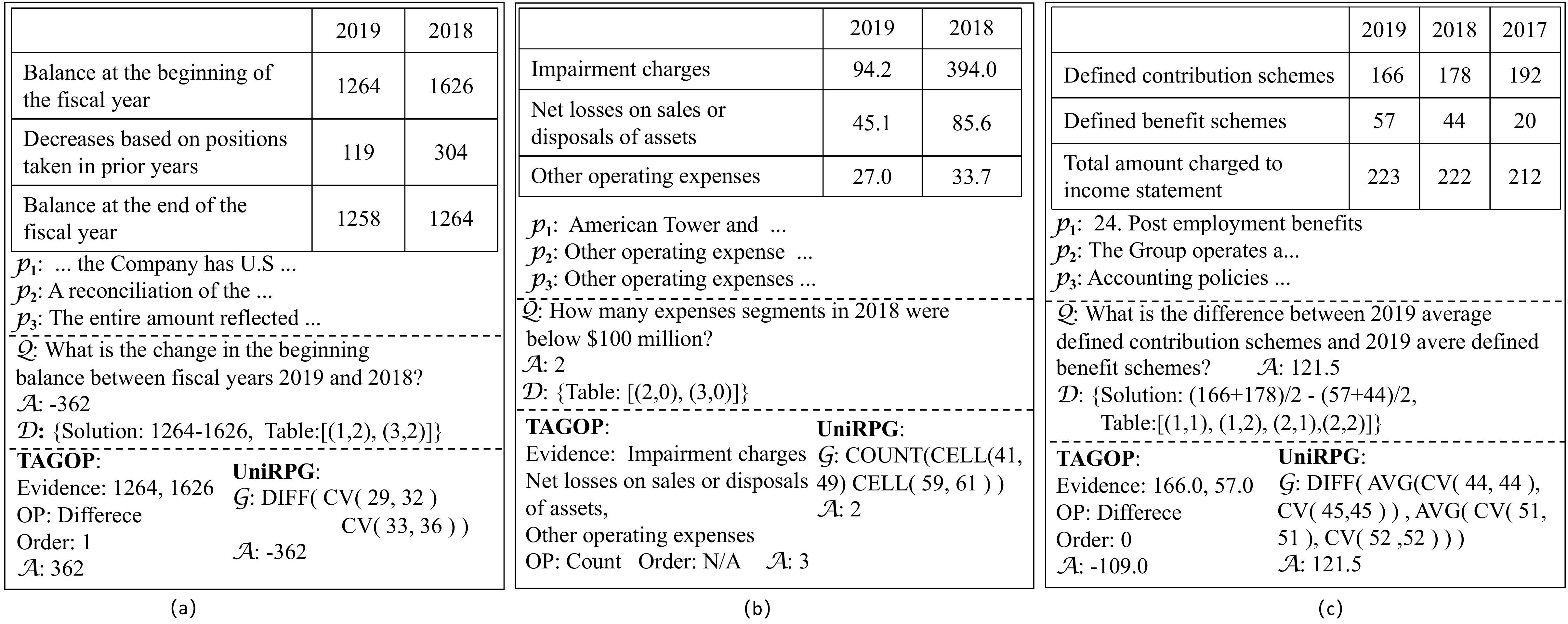}
    \caption{Three examples display the prediction results of baseline TAGOP and our model.
In TAGOP, order=N/A means that the current operation does not require an order, and order is 0 implies that the order of the two numbers is the same as the order in the input sequence; otherwise, it is 1.}
    \label{fig:case}
\end{figure*}

\section{Related Work}
\paragraph{Hybrid QA}
Hybrid QA aims to derive answers to given questions with heterogeneous knowledge, such as table/knowledge graph and text.
Many challenging datasets and promise approaches~\cite{asadifar2018hcqa,chen2020open,chen2020hybridqa,zhu2021tat, zhong2022reasoning,wang2022muger} are recently proposed to stimulate the progress of this topic.
\citet{chen2020hybridqa} proposed a hybrid dataset named HybridQA that each question is aligned with a Wikipedia table and multiple paragraphs linked with cells in the table.
\citet{chen2020open} proposed OTT-QA, a large-scale open table-and-text QA dataset that requires retrieval tables and text to derive answers.  
Unlike HybridQA and OTT-QA, which study multi-hop reasoning on table and text, we study TAT-QA~\cite{zhu2021tat}, a large-scale QA dataset requiring discrete reasoning over hybrid data. 
FinQA~\cite{chen2021finqa} is a recent work similar to ours, aiming to address the task of numerical reasoning over financial data.
Compared to it, UniRPG supports more reasoning capabilities, such as $\mathtt{SPAN}$/$\mathtt{CELL}$, which can perform extraction from paragraphs/tables but is not limited to the arithmetic computing problem.
This enables UniRPG to be available for more discrete reasoning tasks. 
Moreover, the atomic and higher-order operation framework provides a unified paradigm to define operations, which makes UniRPG convenient to be adapted to other discrete reasoning tasks.
In addition, we provide a weak-supervised approach for discrete reasoning over hybrid context based on question-answer pairs, largely reducing the cost of program annotations.

\paragraph{Discrete reasoning for QA}
\citet{dua2019drop} proposed DROP, a large-scale RC dataset to investigate discrete reasoning over text.
Many customized models are proposed to tackle this task, and these fall into two dominant directions,
the first direction is to devise specific modules for each types of question, such as NAQANET~\cite{dua2019drop}, NumNet~\cite{ran2019numnet}, QDGAT~\cite{chen2020question}, EviDR~\cite{zhou2021evidr} and OPERA~\cite{Zhou2022OPERAOD}.
The second direction is to predict programs that would solve the questions~\cite{andor2019giving, gupta2019neural, chen2019neural}.
It is similar to ours, but our model supports discrete reasoning over hybrid data.

\section{Conclusion }
In this work, we propose to perform discrete reasoning over table and text as program generation.
Our framework consists of a neural programmer to parse questions into executable programs and a symbolic program executor to derive answers based on decoded programs. 
Furthermore, we design a distant supervision method to automatically construct pseudo programs to alleviate the cost of manual annotation.
Comprehensive experiments on the TAT-QA and DROP datasets demonstrate the effectiveness and generality of our model under both full and distant supervision.

\section*{Limitations}
In this section, we discuss the limitations of this work.
First, we heuristically search for possible programs based on manually-designed patterns under weak supervision.
It may require expanding and modifying the patterns to derive possible programs when adapted to other tasks.
However, the cost of expanding and modifying the patterns is much less than manually labeling programs(derivation).
Second, we consider all the possible pseudo programs for each instance and
ignore explicitly modeling to choose the correct one from noisy pseudo programs. 
Experiment results show that the distantly supervised model based on automatically constructed pseudo-programs only lags behind fully supervised model 1.0 EM and 1.4 F1 on the test set.
In our experiments, modeling selection of the correct programs from noisy pseudo programs has limited room for improvement, but it remains a valuable topic to explore.

\section*{Acknowledge}
This work is supported by the National Key Research and Development Program of China (No. 2020AAA0108600) and the project of the National Natural Science Foundation of China (No.U1908216).

\normalem
\bibliography{anthology}
\bibliographystyle{acl_natbib}

\clearpage
\appendix
\section{Appendix}

\subsection{Implementation Details}
\label{app:implement details}

In our experiments, we employ BART~\cite{lewis2020bart} as the initialization parameters of UniRPG and optimize it with AdamW~\cite{loshchilov2018decoupled}.
Table~\ref{table:hyper} displays all the hyperparameters for the TAT-QA and DROP datasets during training and inference.
Our experiments are all performed on 1 NVIDIA A100 GPU.

\begin{table}[ht]
  \small
  \centering
  \resizebox{0.48 \textwidth}{!}{
  \setlength{\tabcolsep}{1mm}{
	\begin{tabular}{lccc}
		\toprule
		  \multirow{2}*{\bf Parameters} &\multicolumn{2}{c}{\bf TAT-QA } &   \bf DROP \\
		   \cmidrule(lr){2-3} \cmidrule(lr){4-4}
		   & \bf base & \bf large & \bf large \\
		 \midrule
		  Learning Rate & 1e-4 & 1e-4 &1e-4 \\
		  Batch Size & 128 & 256 & 256 \\
		  Weight Decay & 0.01 & 0.01 & 0.01 \\
		  Gradient Clipping & 5 & 5 & 5 \\
          Warmup Steps & 0.01 & 0.01 & 0.01\\
          Warmup & Linear & Linear & Linear\\
          Maximum Decoding Steps & 50 & 50 & 50\\
          Beam Size & 4 & 4 & 4\\ 
          Training Epochs & 30 & 30 & 20\\
          Lower Layers & 3 & 4 & -\\
          $\lambda$ & 0.3 & 0.3 & -\\
			\bottomrule
	\end{tabular}
	} 
	}
	\caption{\label{table:hyper} Hyperparameters of UniRPG during training and inference on the TAT-QA and DROP dataset.
	}
	\vspace{-0.3cm}
\end{table}

\subsection{Decoding Constraints}
\label{sec:decoding_constraints}
Table \ref{table:decoding_constrains} displays three categories of decoding constraints to ensure program legality, including index, type, and compositional constraints. 
We implement these constraints by proposing a decoding mask to filter illegal program units at each step.  

\begin{table}[ht]
  \small
  \centering
  \setlength\tabcolsep{10pt}
  \resizebox{0.48\textwidth}{!}{
\begin{tabular}{m{0.48\textwidth}}
 \toprule
 
 \bf Constraint Type / Constraints  \\ \midrule
\textit{Index Constraints} \\\tabincell{l}{
\# The end index must equal to or be larger than the start index.\\
\# A $\mathtt{SPAN}$ answer satisfies a length limitation constrained by the \\ ~~~difference between the end and start index. \\
\# We constrain the index to ensure those span items in a \\~~~$\mathtt{MULTI\_SPANS}$ answer are different.
 }
\\ \midrule

\textit{Type Constraints} \\ \tabincell{l}{
\# $\mathtt{CELL}$, $\mathtt{CELL\_VALUE}$, $\mathtt{SPAN}$ and $\mathtt{VALUE}$ should return a correct \\ ~~~type of 
the result, such as $\mathtt{CELL\_VALUE}$ should return a number \\ ~~~from table 
rather than paragraphs.
}\\ \midrule

\textit{Compositional Constraints} \\ \tabincell{l}{
\# The first decoded program unit (except the start identifier $\langle s \rangle$)  \\~~~should be
an atomic operation or higher-order operation,\\~~~ except $\mathtt{KV}$. \\
\# Operations must take correct types of operation as arguments  \\ ~~~that
is exhaustively demonstrated in Table \ref{table:Operations}. \\
\# Set a maximum number of operation augments, such as \\
 ~~~no more than three numbers are involved in $\mathtt{AVG}$.
}\\

\bottomrule
\end{tabular} }
   \centering
   \caption{\label{table:decoding_constrains} Three types of decoding constraints. Each item starting with the character \# means a kind of constraint.}
\end{table}

\subsection{Templates for searching pseudo programs}
\label{app:templates}
Table~\ref{table:templates} displays the template library for searching pseudo programs under the w/o derivation setting.
$\mathtt{F}_1$ and $\mathtt{F}_2$ is anyone of $\{\mathtt{SUM}, \mathtt{DIFF}, \mathtt{TIMES}, \mathtt{DIV}\}$. 
$\mathtt{CV}$ is the abbreviation of operation $\mathtt{CELL\_VALUE}$.
$\mathtt{C}$ means a constant defined in Table~\ref{table:Operations}.
Note that each operation with a numeric subscript means an instance of this operation.

\begin{table}[ht]
  \small
  \centering
  \setlength\tabcolsep{10pt}
  \resizebox{0.48\textwidth}{!}{
\begin{tabular}{m{0.48\textwidth}}
 \toprule
 
 \bf Categories / Program Templates  \\ \midrule

\textit{Text/Number Extraction} \\ \tabincell{l}{$\mathtt{CELL(s,e)}$ \\ $\mathtt{CV(s,e)}$ \\ $\mathtt{SPAN(s,e)}$\\ $\mathtt{VALUE(s,e)}$ } \\ \midrule

\textit{Arithmetic computing} \\ \tabincell{l}{
$\mathtt{AVG}(\mathtt{CV}_1/\mathtt{VALUE}_1, \mathtt{CV}_2/\mathtt{VALUE}_2)$ \\
$\mathtt{AVG}(\mathtt{CV}_1/\mathtt{VALUE}_1, \mathtt{CV}_2/\mathtt{VALUE}_2, \mathtt{CV}_3/\mathtt{VALUE}_3)$ \\
$\mathtt{DIFF}(\mathtt{AVG}_1, \mathtt{AVG}_2)$ \\
$\mathtt{CHANGE\_R}(\mathtt{CV}_1/\mathtt{VALUE}_1, \mathtt{CV}_2/\mathtt{VALUE}_2)$ \\
$\mathtt{DIFF}(\mathtt{CHANGE\_R}_1, \mathtt{CHANGE\_R}_2)$ \\
$\mathtt{F}_1(\mathtt{CV}_1/\mathtt{VALUE}_1/\mathtt{CONST}, \mathtt{CV}_2/\mathtt{VALUE}_2/\mathtt{CONST})$\\
$\mathtt{F}_1(\mathtt{F}_2(\mathtt{CV}_1/\mathtt{VALUE}_1/\mathtt{C}, \mathtt{CV}_2/\mathtt{VALUE}_2/\mathtt{C}), \mathtt{CV}_3/\mathtt{VALUE}_3/\mathtt{C}))$\\
} \\ \midrule

\textit{Comparison Sorting} \\ \tabincell{c}{$\mathtt{ARGMAX}(\mathtt{KV}_1, ...,\mathtt{KV}_m)$ \\
    $\mathtt{ARGMIN}(\mathtt{KV}_1, ...,\mathtt{KV}_m)$
}\\
\midrule

\textit{Multi-spans Extraction} \\  \tabincell{c}{$\mathtt{MULTI\_SPANS}(\mathtt{CELL}_1/\mathtt{SPAN}_1, ... ,  \mathtt{CELL}_m/\mathtt{SPAN}_m)$ 
} \\ 
\midrule

\textit{Counting} \\  \tabincell{c}{$\mathtt{COUNT}(\mathtt{CELL}_1/\mathtt{SPAN}_1, ...,\mathtt{CELL}_m/\mathtt{SPAN}_m)$ \\ 
} \\ 
\bottomrule
\end{tabular} }
   \centering
   
\caption{\label{table:templates} Template library $\mathcal{B}$ for searching pseudo programs under the w/o derivation setting.
 }
\end{table}

\subsection{Examples}
\label{App:examples}
\begin{figure*}[ht]
    \centering
    \includegraphics[width=0.98\textwidth]{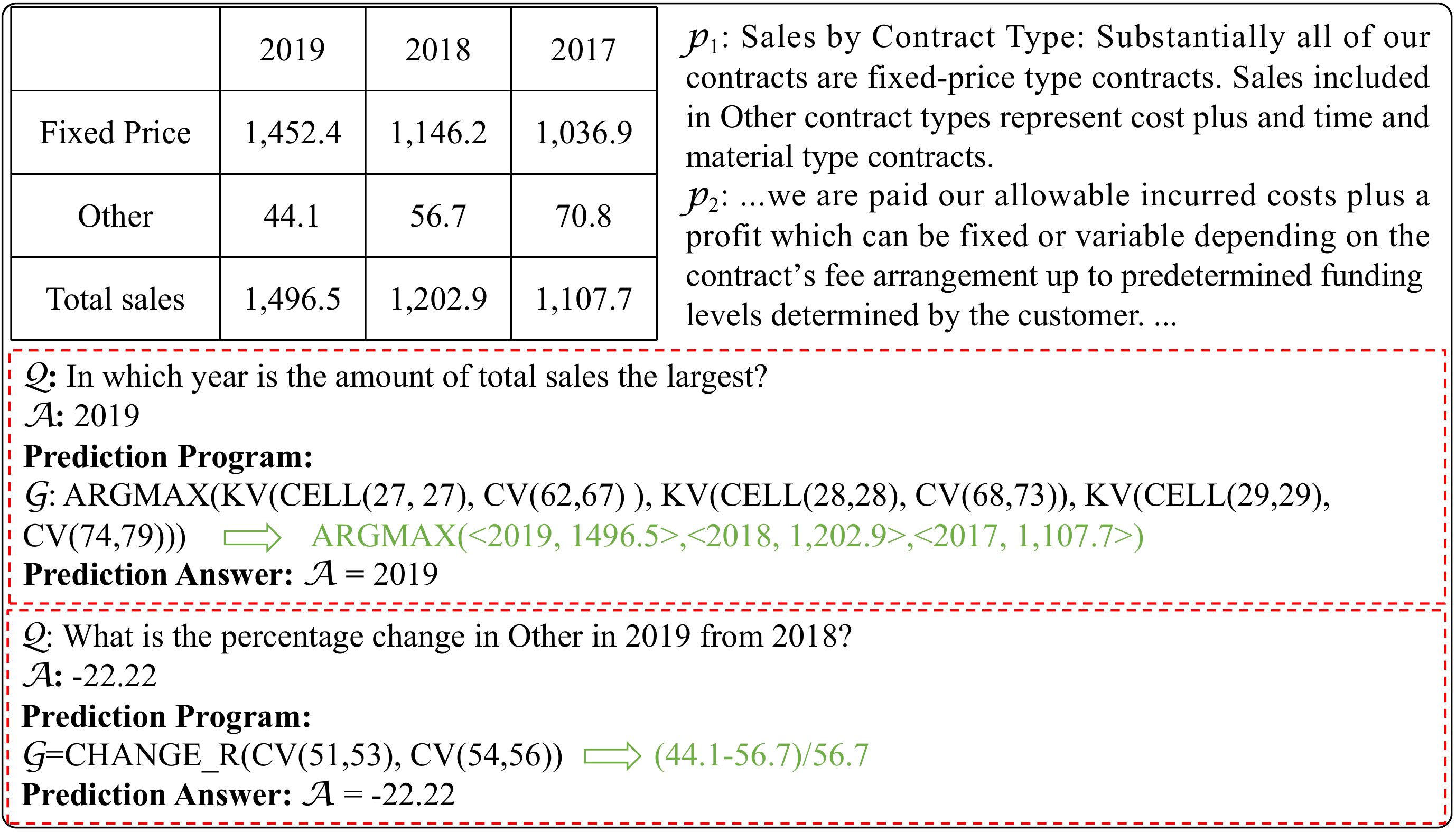}
    \caption{Examples in the development of TAT-QA dataset.}
    \label{Fig:program_examples}
\end{figure*}

\begin{figure*}[ht]
    \centering
    \includegraphics[width=0.98\textwidth]{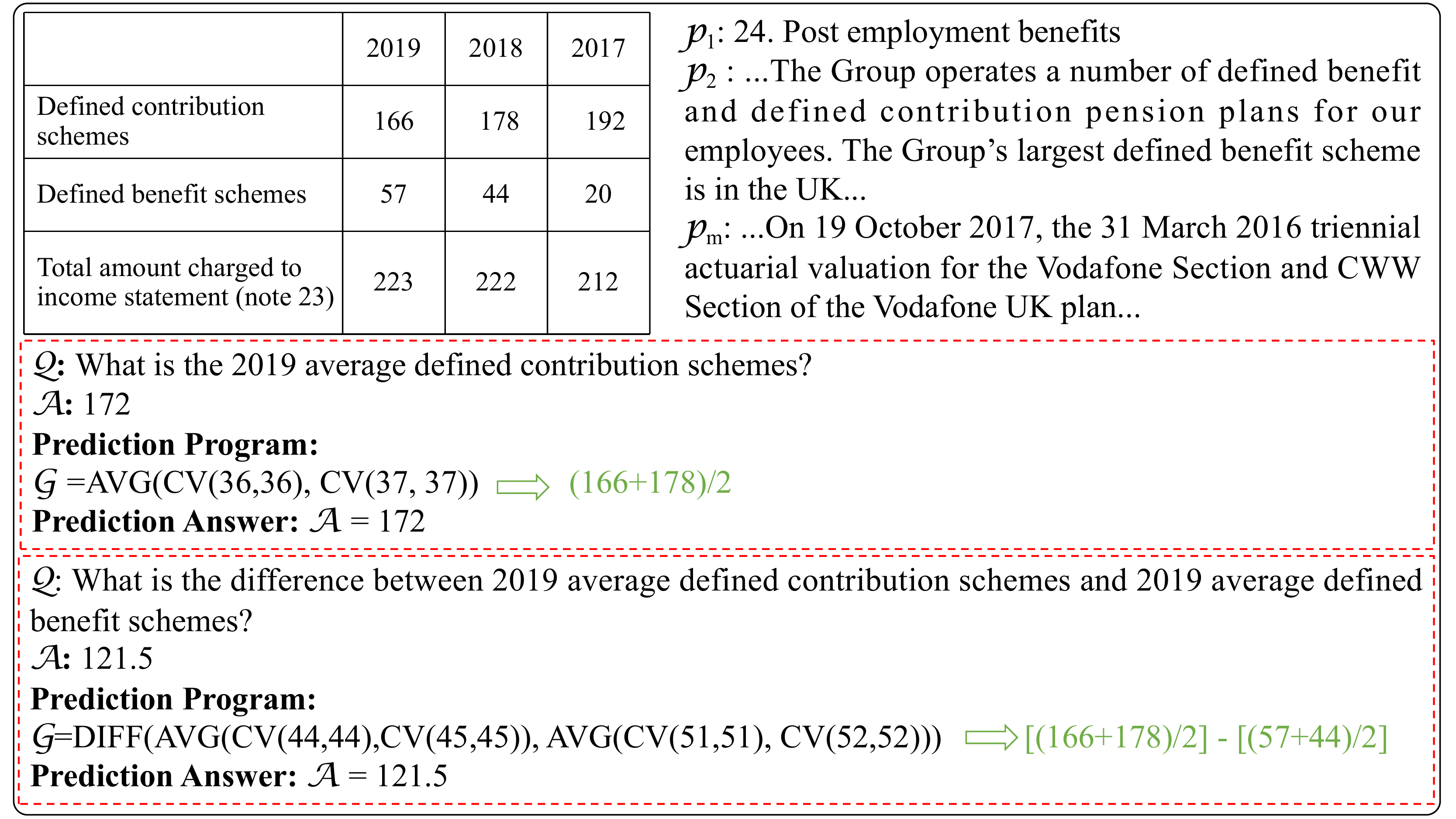}
    \caption{Examples in the development of TAT-QA dataset.}
    \label{Fig:program_examples_2}
\end{figure*}

\begin{figure*}[ht]
    \centering
    \includegraphics[width=0.98\textwidth]{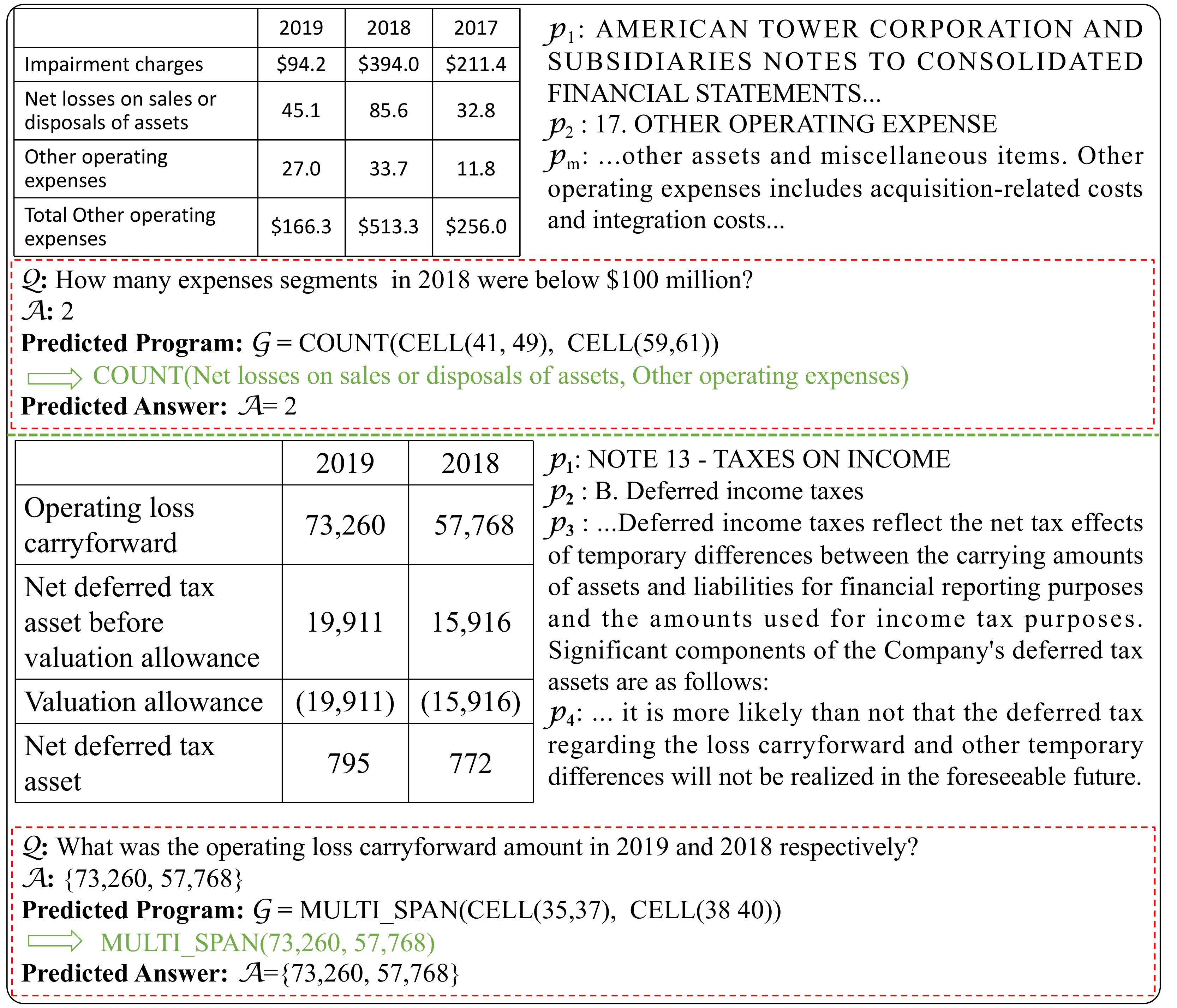}
    \caption{Examples in the development of TAT-QA dataset.}
    \label{Fig:program_examples_3}
\end{figure*}

 Figure~\ref{Fig:program_examples}-~\ref{Fig:program_examples_3} displays several instances, decoded programs and predictions.
Take the first instance in Figure~\ref{Fig:program_examples} as example, UniRPG translates the question ``\textit{In which year is the amount of total sales the largest?}'' as the program 
$\mathcal{G} = \mathtt{ARGMAX(KV(CELL(27, 27), CV(62,67))},  \\ \left. \mathtt{KV(CELL(28, 28), CV(68,73))},  \right. \\\left.  \mathtt{KV(CELL(29, 29), CV(74, 79)))}\right.$.
Instead of only outputting a string ``2019'', we decode the program $\mathcal{G}$ that selects the key with largest value from the key-value pairs $\{\langle 2019, 1496.5 \rangle, \langle 2018, 1202.9 \rangle , \langle 2017, 1107.7 \rangle \}$ to derive the answer.
Intuitively,  our model generates fine-grained programs, illustrating the discrete reasoning process over table and text.

\end{document}